\pdfoutput=1

\documentclass[11pt]{article}

\usepackage[review]{ACL2023}

\usepackage{times}
\usepackage{latexsym}
\usepackage{booktabs}
\usepackage{multirow}
\usepackage{graphicx} 
\usepackage{algorithm}
\usepackage{algorithmic}
\usepackage{amsmath}

\usepackage[T1]{fontenc}

\usepackage[utf8]{inputenc}

\usepackage{microtype}

\usepackage{inconsolata}

\title{Epi-Curriculum: Episodic Curriculum Learning for Low-Resource Domain Adaptation in Neural Machine Translation}

\author{Keyu Chen \\ University of South Florida \\  Tampa, FL 33620 \\  \AND
        Di Zhuang \\ Snap Inc. \\  Santa Monica, CA, 90405  \And
        Mingchen Li \\ University of South Florida \\  Tampa, FL 33620  \And 
        J. Morris Chang \\ University of South Florida \\  Tampa, FL 33620 
        }

\begin{document}
\maketitle
\begin{abstract}
Neural Machine Translation (NMT) models have become successful, but their performance remains poor when translating on new domains with a limited number of data. 
In this paper, we present a novel approach Epi-Curriculum to address low-resource domain adaptation (DA), which contains a new episodic training framework along with denoised curriculum learning.
Our episodic training framework enhances the model's robustness to domain shift by episodically exposing the encoder/decoder to an inexperienced decoder/encoder.
The denoised curriculum learning filters the noised data and further improves the model's adaptability by gradually guiding the learning process from easy to more difficult tasks.
Experiments on English-German and English-Romanian translation show that: 
(i) Epi-Curriculum improves both model's robustness and adaptability in seen and unseen domains; (ii) Our episodic training framework enhances the encoder and decoder's robustness to domain shift.
\end{abstract}


\section{Introduction}

Neural Machine Translation (NMT) \cite{sutskever2014sequence} has yielded state-of-the-art translation performance for many tasks \cite{barrault2019findings, lewis-etal-2020-bart, raffel2020exploring}. 
However, the models perform poorly on the domains with very different statistics to the data used to train them \cite{koehn-knowles-2017-six, chu-wang-2018-survey}.
For instance, a model trained exclusively on the news domain is unlikely to have a good performance on the medical domain. 
It has been approved that models can be trained to perform well on the given domains with large in-domain corpora \cite{barrault2019findings, bawden-etal-2019-findings}, but it is always possible to have some new domains with very limited data as previous works stated \cite{zoph2016transfer, gu-etal-2018-meta}.
Thus, we would like to have a model that is robust enough to address domain adaptation with limited data available.

Having a good model for low-resource domain adaptation is challenging.
First, the model should have good adaptability that is able to quickly adapt to a new domain with only hundreds of in-domain corpus.
We also want the model to be robust (e.g., perform well) before fine-tuning because a much worse model is unlikely to perform well after fine-tuning, even if the model is able to adapt quickly \cite{lai-etal-2022-improving-domain}. 
The challenges have generated significant interest in the literature.
Existing works span adding auxiliary networks to be aware of the domain shifting \cite{bapna-firat-2019-simple, lin-2021-corpus}, introducing training curricula to let the model adapt gradually \cite{bengio2009curriculum, wang-etal-2020-learning-multi} and training the model with some special learning schemes to obtain a set of good initialized parameters for quick adaptation. \cite{sharaf-etal-2020-meta, park-etal-2021-unsupervised, zhan2021meta}.
However, these works either only show the model's robustness to domain shift or the superiority in adapting to new domains with a limited number of data \cite{lai-etal-2022-improving-domain}.
A solution for addressing both model's robustness and adaptability is underexplored. 

In this paper, we present Epi-Curriculum, which contains two main components: a novel episodic training framework and curriculum learning.
The episodic training framework is able to improve the ability of a general encoder-decoder-based NMT model to handle domain shift.
Specifically, it trains the model by synthesizing the real domain shift when an encoder/decoder combines with another inexperienced decoder/encoder that has never trained on such domains before. 
By optimizing the combination of an encoder/decoder and an inexperienced decoder/decoder, both the encoder and decoder will be robust enough to overcome domain shift.
The insight is that, a neural network performs poorly in a new domain because the input statistics are different from the network’s expectations \cite{li2019episodic}.
In other words, the current layer accepts unexpected statistics from its previous layer, and the current layer will also produce unexpected statistics for the next layer.
Therefore, if the neural network can be trained to perform well with unexpected inputs, its robustness to new domains will be enhanced.

Curriculum learning imitates the learning order of human education, which ranks the training data from easy to difficult and gradually presents more difficult tasks to the NMT model during training \cite{bengio2009curriculum, moore-lewis-2010-intelligent, wang-etal-2020-learning-multi, zhan2021meta}.
Inspired by this, curriculum learning is plugged into our episodic training framework to guide the model for better adaptation.
We follow the general curriculum learning framework of difficulty measure and training scheduler \cite{wang2021survey}, where difficulty measure determines the relative ``difficulty'' of each data sample and training scheduler decides the sequence of data subsets throughout the training process. 
Additionally, curriculum learning has been approved as an effective method for data cleaning, and many works have shown that the performance can be improved on denoised data \cite{wang-etal-2017-instance, wang-etal-2019-dynamically}.
Thus, in order to let the model focus on the high-relevant in-domain corpus, curriculum learning is also applied to filter the data.

We evaluate our Epi-Curriculum on English-German (En-De) and English-Romanian (En-Ro) translation tasks with 10 and 9 different domains. 
There are 5 seen domains used for training a teacher model, and then individually fine-tuning on the 5 seen and the rest unseen domains.
BLEU score \cite{papineni2002bleu, post2018call} is reported and the experimental results show that Epi-Curriculum improves the model's robustness and adaptability on both seen and unseen domains. 
For instance, it outperforms the baselines by 1.37 - 3.64 on the En-De task and 1.73 - 3.32 on the En-Ro task.
We further demonstrate the model's robustness to domain shift, where they improve the baseline by 2.55 and 2.59 BLEU scores, respectively.
Our contributions mainly lie in three aspects: 
\begin{itemize}
    \item We propose a novel episodic training framework to handle the model's robustness to domain shift. 
    This is the first work that simulates the situation of unexpected statistics during training for domain adaptation in NMT.
    \item Curriculum learning is applied to our training framework. 
    Our curriculum learning not only guides the model from easy to difficult tasks but also denoises the training data.
    \item We evaluate Epi-Curriculum on two language pairs with ten and nine different domains, empirically showing the strength of our proposed approach. 
\end{itemize}

\section{Related Work}


\subsection{The Encoder-Decoder Model and NMT}
The encoder-decoder model has been widely used as the standard architecture for NMT \cite{bahdanau2014neural}. 
Given a source sentence $S = (s_1, s_2, ..., s_N)$, the encoder-decoder model maps it into a target sentence $T = (t_1, t_2, ..., t_M)$, which can be formalized as the product of series of conditional probabilities: 
\begin{equation}
    P(T|S) = \prod_{m=1}^{M}P(t_m|t_{<m}, S)
\end{equation}
Generally, the encoder takes a variable-length source sentence and generates a fixed-length numerical representation, which then directly passes to the decoder and returns a meaningful sentence in the target language.

By introducing the self-attention and multi-head attention to the prevailing encoder-decoder architecture, the transformer model \cite{vaswani2017attention} obtains promising results in many different NMT datasets \cite{barrault2019findings, fraser-2020-findings}.
Following the successes of the transformer model, \citet{raffel2020exploring} extensively propose the T5 model.
Fortunately, T5 has released the model with pre-trained parameters for generality and reproducibility, and we will use the pre-trained T5 model as the test bench in this paper.


\subsection{Domain Adaptation for NMT}
A conventional way for domain adaptation techniques is fine-tuning \cite{luong2015stanford}, which fine-tunes a pre-trained domain-agnostic teacher model and continues training on a small amount of in-domain data to obtain a domain-specific student model.
Most existing DA works are built on two main strategies to obtain a good model for fine-tuning. 
The first strategy is to add trainable parameters to the NMT model.
The auxiliary parameters could be a domain adaptor \cite{bapna-firat-2019-simple} or a new sub-network \cite{pham2019generic, lin-etal-2021-towards}.
Secondly, some training schemes are proposed to improve the performance after fine-tuning: freezing parameters \cite{wuebker-etal-2018-compact, gu2020investigating}, non-MLE training \cite{wang-sennrich-2020-exposure, saunders-byrne-2020-addressing}, and instance weighting \cite{zhang-xiong-2018-sentence, dougal-lonsdale-2020-improving}.

Besides, it is worth noting that curriculum learning is good at improving domain adaptation performance \cite{zhang2018empirical, wang-etal-2021-multi}.
Rather than sampling the training data randomly, it hypothesizes that neural network training can benefit from learning the data in an easy to difficult order \cite{bengio2009curriculum, zhang-etal-2017-boosting}. 
Moreover, many recent works have shown that domain adaptation performance can be further improved by filtering the less relevant or irrelevant training samples \cite{shu2019transferable, zhang-etal-2019-curriculum}, especially in neural machine translation \cite{kumar-etal-2019-reinforcement, wang-etal-2018-denoising, wang-etal-2019-dynamically}. 
In this paper, we follow \citet{wang-etal-2021-multi} to denoise the data, follow \citet{moore-lewis-2010-intelligent} to measure the sentence-level difficulty and design a difficulty-based training scheduler to guide the training.



\subsection{Meta-Learning for NMT}
Meta-learning, also known as learning-to-learn, which has drawn much attention recently \cite{finn2017model, nichol2018first}, in particularly MAML \cite{finn2017model}.
The main idea of MAML is the episodic training strategy, which mimics the train-test behavior within each episode, such that the model can be adapted to new tasks in a few iterations.
As an effective method for handling low-resource scenarios, many works bring this idea to address domain adaptation in NMT \cite{sharaf-etal-2020-meta, li2020metamt, zhan2021meta}.
Although these works inherit MAML's superiority in adaptability, their robustness is not guaranteed.
Inspired by \citet{li2019episodic}, we also propose an episodic training framework in this work, but with a completely different episode construction to improve the model's robustness to domain shift. 


\begin{figure*}[t]
\centering
\includegraphics[width=2.1\columnwidth]{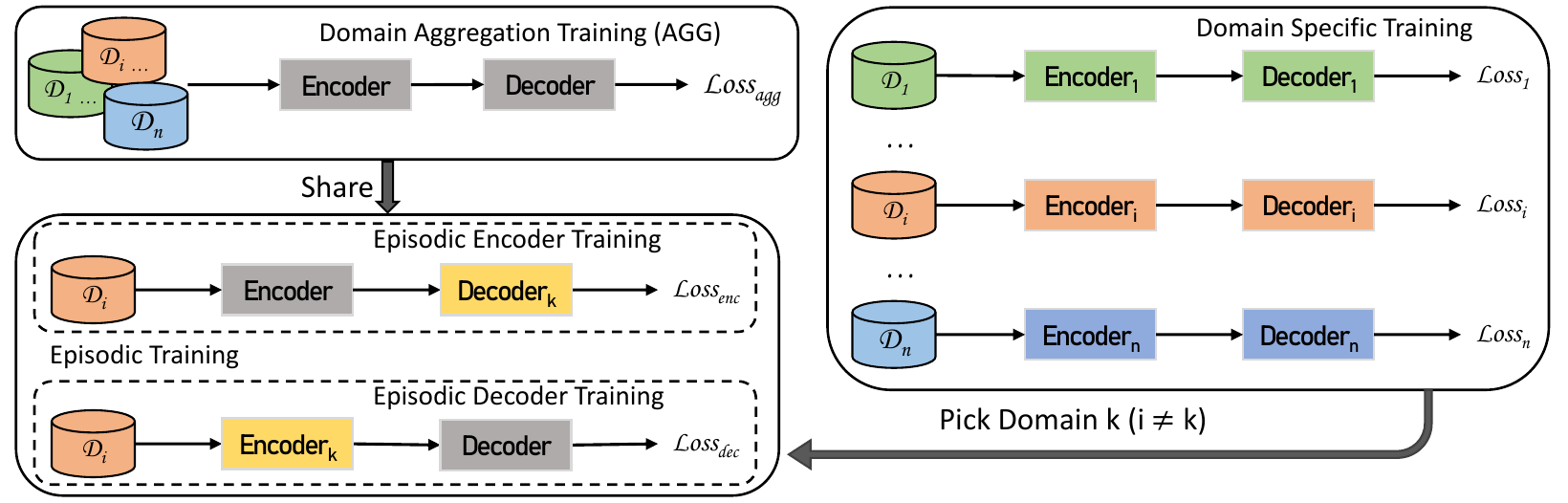} 
\caption{The illustration of our episodic training framework. Four sub-tasks are represented in the diagram, where the Domain Aggregation Training is illustrated on the upper left, the Domain-Specific Training is illustrated on the right and the Episodic Encoder/Decoder Training is illustrated on the bottom left. Best viewed in color.}
\label{all_training}
\end{figure*}

\section{Methodology}

In this section, we first formalize the problem setting of domain adaptation, then subsequently explain the sub-tasks in our episodic framework, followed by our pre-defined curriculum learning.

\subsection{Problem Setting} 
Given $n$ source domains $\mathcal{D} = \{\mathcal{D}_1, ..., \mathcal{D}_n\}$, $\mathcal{D}_{i}$ is the $i^{th}$ source domain containing source ($s$) and target ($t$) sentence pairs $(s^j_i, t^j_i)$, where $j$ is the instance number within domain $i$.
The goal is to train a teacher model on the source domains and then fine-tune the model on a new domain to obtain a student model.
In order to build the encoder/decoder and inexperienced decoder/encoder combination in our episodic framework, we split the encoder-decoder-based NMT model into two modules: an encoder $g$ with parameters $\theta$ and a decoder $h$ with parameters $\phi$.
Thus, for an NMT model $f$ with all parameters $\Theta$, it can be formulated as $f_{\Theta}(s) = h_{\phi}(g_{\theta}(s))$.


\subsection{Episodic Training Framework} \label{agg_spec}
There are four sub-tasks within each episode of our episodic training framework: 
(i) Domain Aggregation Training trains the backbone model.
(ii) Domain-Specific Training provides the inexperienced models.
(iii) Episodic Encoder Training and (iv) Episodic Decoder Training combine the encoder/decoder of the backbone model and the decoder/encoder of an inexperienced model to simulate the domain shift.
The overall training diagram is shown in Figure \ref{all_training}.

\subsubsection{Domain Aggregation Training} 
One traditional domain adaptation pipeline is to continuously train the pre-trained model on the aggregation of all the source domains $\mathcal{D}$ \cite{zoph2016transfer}.
As illustrated in the upper left of Figure \ref{all_training}, both the encoder and decoder will learn from all the domains. 
The optimization is as follows:
\begin{equation}
    \underset{\theta, \phi}{argmin}\sum{_{(s^j_i,t^j_i)\in D}\mathcal{L}_{agg}(t^j_i,h_{\phi}(g_{\theta}(s^j_i)))}
    \label{agg-equation}
\end{equation}
where the $\mathcal{L}_{agg}$ is cross-entropy loss.

Among most of the existing works, this simple approach always performs competitively, and even better than the published methods in some cases \cite{sharaf-etal-2020-meta, zhan2021meta}.
Hence, we use the domain aggregation model (AGG) as the backbone approach of our episodic training framework, and apply the other sub-tasks to improve its robustness throughout the training process.

\subsubsection{Domain-Specific Training}
We improve the robustness by exposing each module (encoder and decoder) of the AGG model to an inexperienced partner.
An inexperienced partner can be an encoder or decoder that has never trained on the corpus of such domains before.
Thus, we employ domain-specific training (right side of Figure \ref{all_training}) to provide the inexperienced partner.
To formalize, for each domain-specific model with its own encoder $g_{\theta _i} $ and decoder $h_{\phi _i}$, we optimize the model using its associated domain corpus $\mathcal{D}_i\in \mathcal{D}$:

\begin{equation}
    \underset{\theta _i, \phi _i}{argmin}\sum{_{(s^j_i,t^j_i)\in D_i}\mathcal{L}_{i}((t^j_i,h_{\phi _i}(g_{\theta _i}(s^j_i)))}
    \label{spec-equation}
\end{equation}
where $\mathcal{L}_{i}$ is the loss of $i^{th}$ domain.
The domain-specific model is only trained on one single domain $i$ and easily performs badly on the other domains.


\subsubsection{Episodic Encoder Training}
To make the encoder robust enough, we consider it should perform well with a decoder that has never trained on such domains before.
As illustrated in the bottom left of Figure \ref{all_training}, given the source domain $i$ and the AGG model, its decoder is replaced by a random decoder $h_{\phi _k} $ of domain-specific models. 
Then the encoder and the inexperienced decoder will together try to translate the sentence in the domain $i$ and compute episodic encoder loss $\mathcal{L}_{enc}$.
We formalize the optimization as follows:

\begin{equation}
    \underset{\theta}{argmin}\sum{_{(s^j_i,t^j_i)\in D_{i\neq k}}\mathcal{L}_{enc}((t^j_i,h_{\bar{\phi} _k}(g_{\theta }(s^j_i)))}
\end{equation}
where $i \neq k$ is to guarantee that the decoder $h_{\phi _k} $ has no experience with domain $i$.
Note that $h_{\bar{\phi} _k}$ means the parameters $\phi _k$ will not be updated during back-propagation because we want the decoder to remain ignorant about the domains outside of $k$.  
Intuitively, this combination can perform poorly due to the decoder's ignorance.
But by minimizing $\mathcal{L}_{enc}$, the encoder will be trained to encode the source sentences into features that can be decoded correctly by a decoder with no experience. 


\subsubsection{Episodic Decoder Training}
Similarly, to train a robust decoder, we assume that the decoder should be able to correctly decode the features generated by an encoder that has never been exposed to such domains before.
As illustrated in the bottom left of Figure \ref{all_training}, the encoder of the aggregation model is replaced by a random encoder $g_{\theta _k} $ of the domain-specific models.
We then ask them to translate the source language in the domain $i$ and compute episodic decoder loss $\mathcal{L}_{dec}$. 
To minimize the loss, we optimize:
\begin{equation}
    \underset{\phi}{argmin}\sum{_{(s^j_i,t^j_i)\in D_{i\neq k}}\mathcal{L}_{dec}((t^j_i,h_{\phi}(g_{\bar{\theta}_{k}}(s^j_i)))}
\end{equation}
where $i \neq k$ is to ensure that $i$ mismatches $k$. 
Similar to the optimization policy of episodic encoder training, $g_{\bar{\theta} _k}$ aims to keep the parameters $\theta _k$ unchanged and to remain ignorant about domains other than $k$. 
Eventually, by minimizing the loss $\mathcal{L}_{dec}$, the decoder of AGG gradually learns to correctly decode the features generated by a random and inexperienced encoder.


\subsection{Pre-defined Curriculum Learning}
There are two main factors in our curriculum learning: 
(i) Data Denoising to filter the corpus.
(ii) Divergence Scoring to measure the sentence-level difficulty of each sample.

\subsubsection{Data Denoising}
\label{sec_denoise}
Given a sentence pair $(s, t)$ of domain $Z$, we follow \citet{wang-etal-2021-multi} to evaluate its cross-entropy difference between two NMT models:

\begin{equation}
\label{denoise_equation}
    q_Z(s,t) = \frac{logP(t|s;\Theta_Z) - logP(t|s;\Theta_{base})}{|t|}
\end{equation}
$P(t|s;\Theta_{base})$ is the base model with parameters $\Theta_{base}$ trained on general domains. 
$P(t|s;\Theta_Z)$ is a domain-specific model with parameters $\Theta_Z$ by fine-tuning the base model on a small $Z$-domain parallel corpus $\widehat{D}_Z$ with trusted quality.
The cross-entropy difference $q_Z$ is normalized by the target sentence length $|t|$.
Additionally, \citet{grangier-iter-2022-trade} has shown that the $q_Z$ with positive value has a positive influence in gradient for adapting a base model to domain $Z$.
Thus, for a multi-domain adaptation problem, the model benefits from a batch of samples with all positive values.
In this work, we filter the samples with negative $q_Z$ because we consider those samples will have the opposite influence on domain adaptation.

\subsubsection{Divergence Scoring}
\label{sec_divergence}
For a given sentence $s$, we follow \citet{moore-lewis-2010-intelligent} to evaluate its cross-entropy difference between two neural language models (NLM) as its $Z$-domain divergence score $d_Z(s)$:
\begin{equation}
\label{divergence_score}
    d_Z(s) = \frac{logP(s;\bar{\Theta}_Z) - logP(s;\bar{\Theta}_{base})}{|s|} 
\end{equation}
$P(s;\bar{\Theta}_{base})$ is the base language model with parameters $\bar{\Theta}_{base}$ trained on general domains.
While $P(s;\bar{\Theta}_Z)$ is the $Z$-domain language model obtained by fine-tuning the base model on $Z$-domain monolingual data.
The $d_Z(s)$ is normalized by the sentence length of $s$.
The higher divergence score indicates that the given sentence $s$ is more divergent from the samples in the general domain, so we will present it to the model in the later stage.


\subsection{Overall Training Flow}
To summarize, our full Epi-Curriculum contains three main steps:
(i) We first use Equation \ref{denoise_equation} to denoise the training corpus by filtering the samples with negative $q_Z$ values.
(ii) Secondly, the rest corpus is sorted according to Equation \ref{divergence_score}.
(iii) Once the data are properly processed, a divergence-score-based training scheduler is plugged into our episodic training framework.
The full pseudocode of our Epi-Curriculum training policy is given in Algorithm \ref{epi-algorithm}, where $\alpha$ and $\beta$ indicate the learning rate of aggregation optimization and domain-specific optimization, respectively.

\begin{algorithm}[tb]
\caption{Epi-Curriculum Training Policy}
\label{epi-algorithm}
\textbf{Require}: Aggregation Model: ($g_{\theta} $, $h_{\phi} $);
Domain-Specific Models: \{($g_{\theta_1}$, $h_{\phi_1}$), ..., ($g_{\theta_n} $, $h_{\phi_n}$)\};
Pre-trained NMT Models: \{$\Theta_1, ...,\Theta_n$\};
Pre-trained NLM Models: \{$\bar{\Theta}_1$, ... $\bar{\Theta}_n$\};
Base NMT model: $\Theta_{base}$;
Base NLM model: $\bar{\Theta}_{base}$
\\
\textbf{Hyperparameters}: $\alpha, \beta$\\
\textbf{Output}: ($g_{\theta} $, $h_{\phi}$)
\begin{algorithmic}[1] 
\STATE Use $\Theta_{base}$ and $\Theta_{Z}$ to score the $q_Z$ of each sentence in $\mathcal{D}$
\STATE Filter the sentences with negative $q_Z$
\STATE Use $\bar{\Theta}_{base}$ and $\bar{\Theta}_{Z}$ to score the $d_Z$ and sort the rest sentences
\WHILE{not done}
\STATE Sample the sentences based on the sorted $d_Z$ with pre-defined probabilities
\FOR{($g_{\theta_i}$, $h_{\phi_i}$)$\in$\{($g_{\theta_1}$, $h_{\phi_1}$), ..., ($g_{\theta_n}$, $h_{\phi_n} $)\}}
\STATE Update $\theta _i$ $\gets$ $\theta_i$ - $\beta \nabla _{g_{\theta_i}}\mathcal{L}_i$
\STATE Update $\phi _i$ $\gets$ $\phi_i$ - $\beta \nabla _{h_{\phi_i}}\mathcal{L}_i$
\ENDFOR
\STATE Update $\theta$ $\gets$ $\theta$ - $\alpha \nabla _{g_{\theta}}(\mathcal{L}_{agg} + \mathcal{L}_{enc})$
\STATE Update $\phi$ $\gets$ $\phi$ - $\alpha \nabla _{h_{\phi}}(\mathcal{L}_{agg} + \mathcal{L}_{dec})$
\ENDWHILE
\end{algorithmic}
\end{algorithm} 


\section{Experiments}
In this section, the experiments are designed to investigate the following questions:
(i) How does our Epi-Curriculum empirically compare to the baselines and alternative domain approaches?
(ii) Do our encoder and decoder have the strength to overcome domain shift? 
(iii) What is the impact of other variants of the curriculum?
To explore these, we conducted experiments on English-German (En-De) and English-Romanian (En-Ro) translation tasks with ten and nine different domains.


\begin{table*}
\centering
\resizebox{2\columnwidth}{!}{
\begin{tabular}{c c cccccc c cccccc}
\hline
&  & \multicolumn{5}{c}{\textbf{Unseen}} & &  \multicolumn{5}{c}{\textbf{Seen (Training Domains)}} &  \\
\cmidrule{3-7} \cmidrule{9-13}
& & \textit{Covid-19} & \textit{Bible} & \textit{Books} & \textit{ECB} & \textit{TED2013} & &  \textit{EMEA} & \textit{Tanzil} & \textit{KDE4} & \textit{OpenSub} & \textit{JRC} \\
\hline

 & Vanilla & 25.65 & 10.94 & 10.80 & 31.79 & 26.45 & & 29.02 & 9.83 & 19.67 & 16.25 & 31.87 \\
& AGG & 25.92 & \textbf{12.74} & 10.48 & 32.70 & 26.52 &  &  37.95 & 16.08 & 30.15 & 19.36 & 39.58 \\
Before & AGG-Curriculum & 26.05 & 12.06 & 11.36 & 32.66 & 26.91 & &  38.46 & 15.89 & 30.94 & 18.73 & 40.61 \\
FT & Meta-MT & 25.72  & 12.04 & 10.61 & 32.37 & 26.17 &  & 37.46 & 16.27 & 29.82 & 18.77 & 38.76\\
& Epi-NMT & 26.63 & 12.51 & 11.65 & 32.96 & \textbf{26.93} & &  42.13 & 17.79 & 32.15 & \textbf{20.57} & 40.24 \\
& Epi-Curriculum & \textbf{27.12} & 12.70 & \textbf{11.90} & \textbf{34.12} & 26.63 & &  \textbf{43.51} & \textbf{18.72} & \textbf{32.87} & 20.44 & \textbf{42.30} \\
\hline

& Vanilla & 25.84 & 11.47 & 11.16 & 32.83 & 26.68 & & 29.32 & 10.25 & 20.97 & 17.03 & 32.15  \\
& AGG & 26.39 & 13.43 & 11.30 & 33.02 & 27.04 &  & 38.34 & 16.84 & 30.21 & 19.20 & 39.52 \\ 
After & AGG-Curriculum & 26.71 & 13.48 & 11.77 & 33.76 & 28.36 & & 39.40 & 16.78 & 31.24 & 19.43 & 41.10 \\ 
FT & Meta-MT & 26.64 & 13.40 & 11.23 & 33.56 & 26.91 & &  38.86 & 17.05 & 31.44 & 19.45 & 39.18 \\ 
& Epi-NMT & 27.44 & 14.48 & 12.37 & 34.07 & 28.19 & & 42.45 & 18.39 & 33.22 & \textbf{20.97} & 41.24 \\
& Epi-Curriculum & \textbf{27.73} & \textbf{15.11} & \textbf{12.53} & \textbf{34.89} & \textbf{29.11} & & \textbf{44.18} & \textbf{20.62} & \textbf{33.39} & 20.89 & \textbf{43.24} \\ 
\hline

\multirow{6}{*}{$\triangle$FT} & Vanilla & 0.19 & 0.53 & 0.36 & 1.04 & 0.23 & & 0.30 & 0.42 & 1.30 & \textbf{0.78} & 0.28 \\
& AGG & 0.47 & 0.69 & \textbf{0.82} & 0.32 & 0.52 & & 0.39 & 0.76 & 0.06 & -0.16 & -0.06\\
& AGG-Curriculum & 0.66 & 1.42 & 0.41 & 1.10 & 1.44 & & 0.93 & 0.89 & 0.30 & 0.69 & 0.48\\
& Meta-MT & \textbf{0.92} & 1.36 & 0.62 & \textbf{1.19} & 0.74 && \textbf{1.40} & 0.78 & \textbf{1.62} & 0.68 & 0.42\\
& Epi-NMT & 0.81 & 1.97 & 0.72 & 1.11 & 1.26 & & 0.32 & 0.60 & 1.07 & 0.40 & \textbf{1.00}\\
& Epi-Curriculum & 0.61 & \textbf{2.41} & 0.63 & 0.77 & \textbf{2.48} & & 0.67 & \textbf{1.90} & 0.52 & 0.45 & 0.94 \\
\hline
\end{tabular}
}
\caption{BLEU scores over Testing sets on En-De task. Before/After FT denotes the performance before/after individual fine-tuning. $\triangle$FT denotes the improvement of fine-tuning. The best results are highlighted in bold.}
\label{bleu_score}
\end{table*}

\subsection{Datasets}
The data sources for our two tasks are the following:
\textbf{En-De}: It consists of ten parallel corpora, where 9 of them are collected from the Open Parallel Corpus (OPUS) \cite{tiedemann-2012-parallel} and the rest Covid-19 is from authentic public institution data sources\footnote{\url{https://www.bundesregierung.de/}}$^,$\footnote{\url{https://www.euro.who.int/en/health-topics/health-emergencies/coronavirus-covid-19/news/}}.
These 10 corpora cover a wide range of topics that enable us to evaluate domain adaptation: Covid-19, Bible, Books, ECB, TED2013, EMEA, Tanzil, KDE4, OpenSub and JRC.
\textbf{En-Ro}: All 9 corpora are collected from the Open Parallel Corpus (OPUS): KDE4, Bible, QED, GlobalVoices, EMEA, Tanzil, TED2013, OpenSub, and JRC.

For each translation task, only 5 domains are used for training, and all domains are used for individual fine-tuning evaluation.
Thus, we are able to investigate the model's adaptability on the domains that have never been seen during the training. 
The 5 domains in training are named as \textit{seen} domains and the rest are as \textit{unseen} domains.
Each domain is split into Training (only for seen), Fine-tuning, and Testing.
More details about the data statistics and data preprocessing are shown in Appendix \ref{appendix_data_sta}.


\subsection{Comparison Group} \label{competitors}
The following NMT approaches are included in our experiments:
\textbf{Vanilla}: The system that is pre-trained on the generic domain without adaptation.
It is the default baseline for domain adaptation in NMT after fine-tuning \cite{luong2015stanford}. 
\textbf{AGG (Transfer Learning)}: The domain aggregation model introduced in Equation \ref{agg-equation} without any special training schemes, which continues training on Vanilla with all training corpora. 
It is a strong baseline with comparable performance in many existing works \cite{zhan2021meta, lai-etal-2021-improving}.
\textbf{AGG-Curriculum}: The AGG model trains with our pre-defined curriculum learning strategy.
We are able to observe the sole performance of our curriculum learning.
\textbf{Meta-MT \cite{sharaf-etal-2020-meta}}: The standard meta-training approach that directly applies the MAML algorithm \cite{finn2017model}. 
Including Meta-MT in our comparison is essential as this framework has gained significant popularity in recent works \cite{zhan-etal-2021-augmenting, lai-etal-2021-improving, park-etal-2021-unsupervised}.
We follow the algorithm and implement it on our own for a fair comparison.
\textbf{Epi-NMT}: The approach trains with our episodic framework for evaluating the sole performance of the episodic framework.
\textbf{Epi-Curriculum}: The full version of our approach.


\subsection{Implementation Settings}
T5-small model with pre-trained parameters on C4 (Colossal Clean Crawled Corpus) dataset \cite{raffel2020exploring} is used as the backbone network for all the experiments.
More details about the hyperparameters can be found in Appendix \ref{appendix_hyperparameters}.

To design the training scheduler for our curriculum learning, we filter the noise data and then follow \citet{zhang2018empirical} to sort and group the training samples into five shards evenly, such that those in the same shards have similar divergence scores and the same number of samples. 
More details about the training scheduler of our designed curriculum can be found in Appendix \ref{appendix_curriculum_training_scheduler}.

\subsection{Evaluation}
Once the training is done, we individually adapt each model on the Fine-Tuning set and evaluate its performance on the Testing set.
The SacreBLEU \cite{post2018call} is reported based on the average results from over 5 times training, and the generated results utilize a beam size 5.
There are three types of results we want to observe and highlight: 
\textbf{Before FT:} 
The BLEU score before fine-tuning to demonstrate the model's robustness. 
\textbf{After FT:}
The BLEU score after individual fine-tuning for evaluating the model's final performance.
\textbf{$\Delta$FT:} 
The BLEU improvement through individual fine-tuning for analyzing the model's adaptability.

\begin{table}[t]
\centering
\resizebox{1\columnwidth}{!}{
\begin{tabular}{c | c c | c c }
    \hline
    & \multicolumn{2}{c}{\textbf{Before FT}} & \multicolumn{2}{c}{\textbf{After FT ($\triangle$FT)}}\\
    \hline
     & \textbf{Unseen} & \textbf{Seen} & \textbf{Unseen} & \textbf{Seen} \\
    \hline
    AGG & 0.55 & / & 0.64 (0.09) & / (/)\\
    AGG-Curriculum & 0.68 & 0.30 & 1.22 (0.54) & 0.77 (0.47)\\
    Meta-MT & 0.26 & -0.41 & 0.75 (0.50) & 0.37 (\textbf{0.78})\\
    Epi-NMT & 1.01 & 1.95 & 1.71 (0.70)  & 2.43 (0.48)\\
    Epi-Curriculum & \textbf{1.37} & \textbf{2.94} & \textbf{2.28 (0.91)} & \textbf{3.64} (0.70)\\
    \hline
\end{tabular}
}
\caption{The average improvement of Before FT, After FT, and $\triangle$FT over baselines (Vanilla for unseen and AGG for seen) on the En-De task.}
\label{en_de_small}
\end{table}

\subsection{Results and Discussion (En-De)}
In this section, we discuss the results of the English-German (En-De) task (shown in Table \ref{bleu_score} and Table \ref{en_de_small}).
Table \ref{bleu_score} shows the performance of Before FT, After FT, and $\triangle$FT for each domain.
While Table \ref{en_de_small} shows the average improvement of Before FT, After FT and $\triangle$FT over the baselines.
The baseline for unseen is Vanilla and for seen is AGG, because Vanilla never trained on the seen domains.
The results and discussion of the English-Romanian (En-Ro) task can be found in Appendix \ref{appendix_en_ro}.

\textbf{Robustness:}
Based on the results Before FT in Table \ref{bleu_score}, we can see that:
(i) Compare to Vanilla, AGG shows significant BLEU score gaps in seen domains due to training on the source data. 
(ii) The meta-learning approach Meta-MT is even worse than the AGG in 6 domains.
(iii) Comparing our proposed episodic framework (Epi-NMT) and curriculum learning (AGG-Curriculum) solely, Epi-NMT outperforms AGG-Curriculum in 9 out of 10 domains. 
(iv) Our episodic-based approaches (Epi-NMT and Epi-Curriculum) have the best performance in 9 out of 10 domains.

\textbf{Adaptability:}
From the results of After FT and $\triangle$FT in Table \ref{bleu_score}, we can observe that:
(i) Unlike the performance before fine-tuning, Meta-MT demonstrates its superiority in $\triangle$FT and surpasses AGG after fine-tuning in 6 domains. 
(ii) Epi-Curriculum shows its strength in performance after fine-tuning, where it performs the best in 9 out of 10 domains.

\textbf{Summary:}
Based on the results in Table \ref{en_de_small}, we can observe that: 
(i) Although Meta-MT has strength in adaptability (0.50 and 0.78), its weakness in robustness is also obvious, which has the lowest improvement Before FT.
(ii) Epi-NMT consistently outperforms AGG-Curriculum in robustness and achieves comparable adaptability (0.70 and 0.48), demonstrating the robustness and adaptability of the episodic framework.
(iii) Our proposed episodic framework Epi-NMT is only worse than Meta-MT on $\triangle$FT in seen domains, indicating that the episodic framework is more effective than the MAML-based framework.
(iv) Our full version approach Epi-Curriculum consistently performs the best in Before FT (1.37 and 2.94), After FT (2.28 and 3.64), and $\triangle$FT (0.91 and 0.70), only the adaptability is slightly worse than Meta-MT in seen domains (0.70 vs 0.78).


\begin{figure}[t]
\centering
\includegraphics[width=0.97\columnwidth]{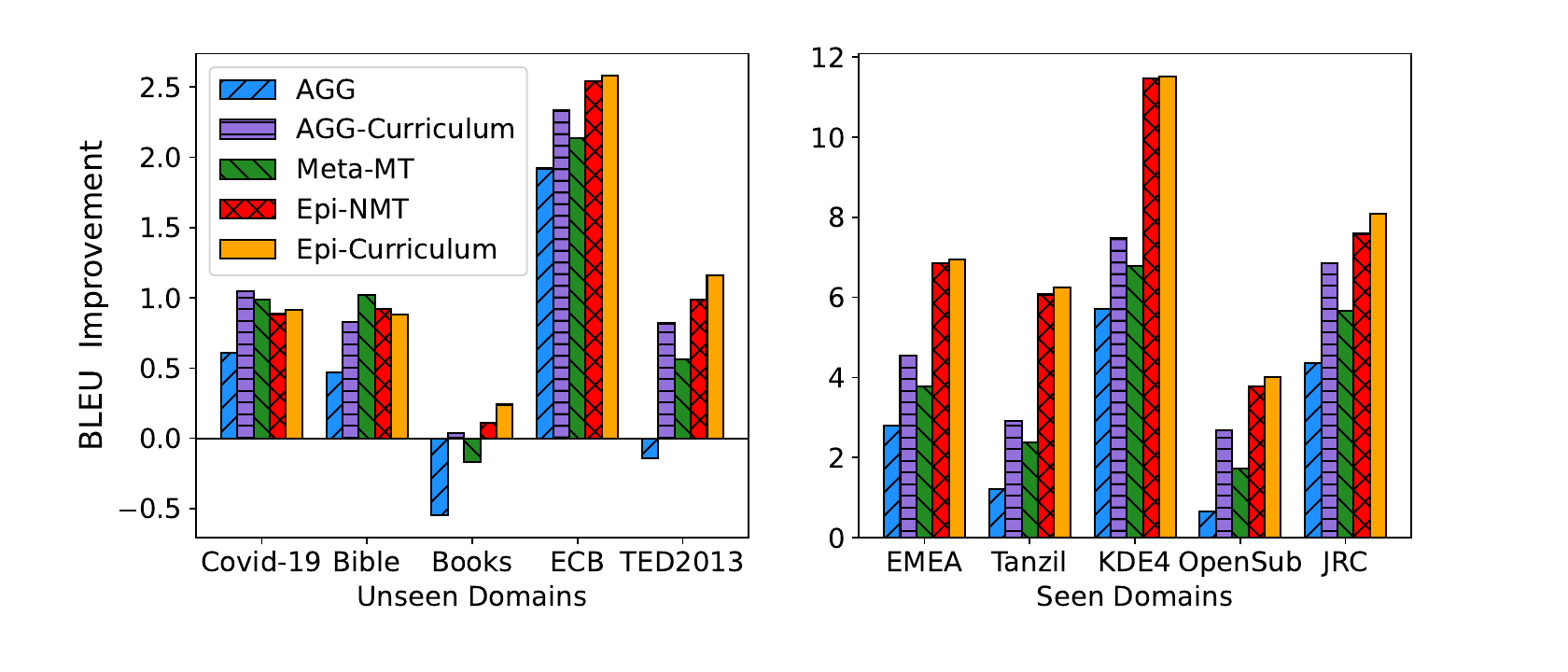} 
\caption{Cross-domain improvement by replacing the encoder of domain-specific models on the En-De task.}
\label{change_encoder}
\end{figure}

\begin{figure}[t]
\centering
\includegraphics[width=0.97\columnwidth]{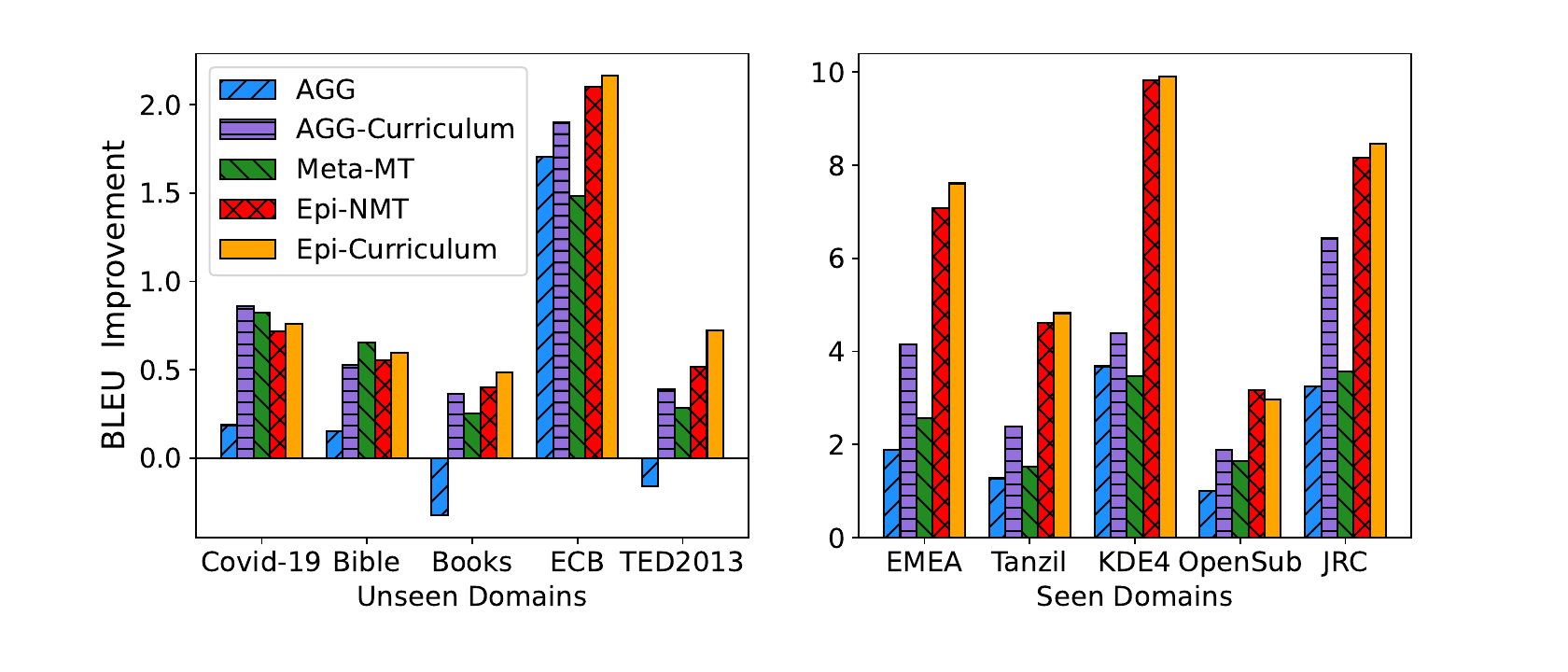} 
\caption{Cross-domain improvement by replacing the decoder of domain-specific models on the En-De task.}
\label{change_decoder}
\end{figure}

\subsection{Robustness to Domain Shift}
To understand how the episodic framework improves the model's robustness to domain shift, we compare its impact on cross-domain testing with other approaches on the En-De task.
The BLEU improvement is reported by evaluating the performance of an encoder/decoder of an inexperienced model, and combined with the decoder/encoder trained by Epi-Curriculum, where the inexperienced model is one of the domain-specific models (introduced in Equation \ref{spec-equation}).
For instance, to compute the encoder improvement on the Covid-19 domain, we replace the encoder of domain-specific models (EMEA, Tanzil, KDE4, OpenSub, and JRC) with the encoder trained by Epi-Curriculum, test them on the Testing set of Covid-19 and report the average BLEU improvements. 
The model is excluded when it matches the input domain to maintain the cross-domain testing.

The results are shown in Figure \ref{change_encoder} and Figure \ref{change_decoder} for encoder and decoder, respectively.
We can see that:
(i) AGG and Meta-MT have negative impacts in the Books and TED2013 domain. 
(ii) AGG-Curriculum has no negative impact and is only slightly lower than Meta-MT in the Bible domain.
(iii) Epi-NMT and Epi-Curriculum perform very close and have the best performance in 8 out of 10 domains, except the Covid-19 and Bible.  
To quantize, Epi-Curriculum outperforms AGG by 2.55 and 2.59 BLEU scores in the case of encoder and decoder, respectively.
This experiment demonstrates that our episodic training strategy indeed enhances the model's robustness to domain shift.

Moreover, we also investigate the robustness in terms of the model's generalization ability.
The results suggest that the model trained by our episodic framework is able to find better local minima.
More details can be found in Appendix \ref{appendix_perturbation}.


\subsection{Curriculum Validity (En-De)}

\begin{figure}[t]
\centering
\includegraphics[width=0.75\columnwidth]{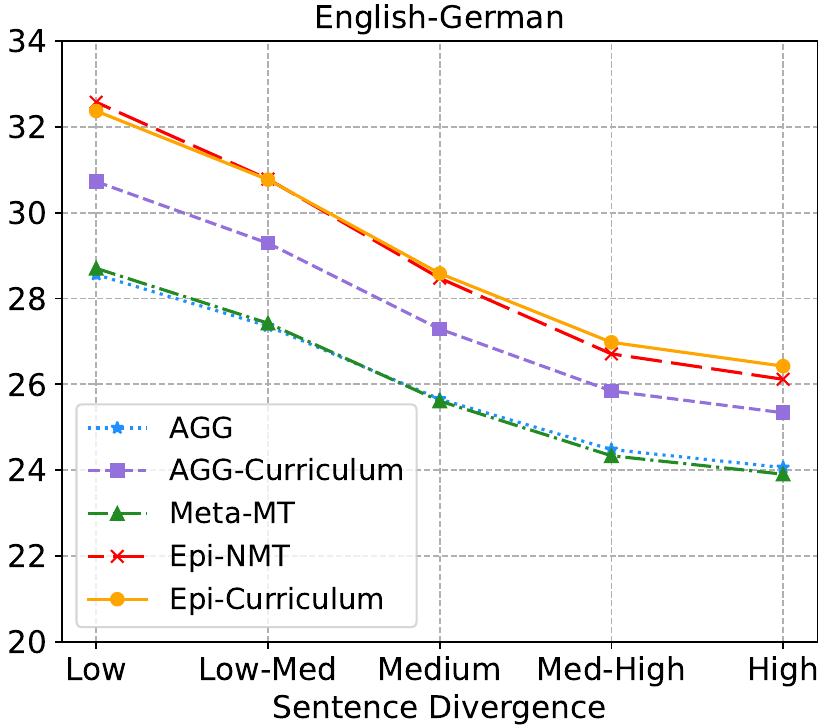} 
\caption{Average performance over Testing sets across five levels of divergence on the seen domains (En-De).}
\label{curr_de}
\end{figure}

To verify our curriculum validity, we categorize the sentences in the Testing set into 5 divergence levels with the thresholds used in the Training set, where each shard has a similar number of samples.
The result of the En-De task is shown in Figure \ref{curr_de}, we can observe that:
(i) The BLEU scores gradually decrease with increasing divergence levels, indicating that the metric introduced in Equation \ref{divergence_score} is able to evaluate adaptation difficulty.
(ii) Epi-NMT and Epi-Curriculum have much better performance in all levels of divergence groups, showing the robustness of our episodic framework. 
(iii) Compared to Epi-NMT, Epi-Curriculum has strength in the shards with higher divergence scores, demonstrating the effectiveness of the designed curriculum.
The result of the En-Ro task has a similar pattern, more details are shown in Appendix \ref{appendix_curr_divergence}.

\subsection{Impact of Training Schedulers}
\label{impact_of_training_schedulers}

The curriculum training scheduler defines the order in which samples of different divergence shards are presented to the training process.
It is natural to introduce the shards from low to high divergence \cite{bengio2009curriculum}.
However, there does not exist golden standard for the training scheduler.

We introduce two variants of training schedulers to investigate the impact of different schedulers on curriculum-based approaches (AGG-Curriculum and Epi-Curriculum). 
The two variants and the experimental results can be found in Appendix \ref{appendix_curriculum_training_scheduler}.
Same as \citet{zhang2018empirical}, our results indicate that curriculum learning can lead to better results, but the default low-to-high divergence curriculum is not the only useful curriculum strategy.

\subsection{Impact of Denoising}
To understand the impact of data denoising in our approach, we compare the performance of AGG-Curriculum and Epi-Curriculum trained with noise data.
Specifically, only Equation \ref{divergence_score} is applied to sort the data and without using Equation \ref{denoise_equation} to filter them.
For both En-De and En-Ro tasks, the results are very close, and hard to conclude which one is better.
This is because it is difficult to affect a translation model with only approximately 8\% less amount of corpus.
But it also proves the effectiveness of our approach, which can achieve the same performance with 8\% less amount of data.
Experimental details can be found in Appendix \ref{appendix_denoise_impact}.


\section{Conclusion}
We present Epi-Curriculum for low-resource domain adaptation in NMT. 
A novel episodic framework is proposed to handle the model's robustness to domain shift, and a denoised curriculum learning is applied to further boost the model's adaptability.
Experiments on En-De and En-Ro empirically show the effectiveness of our approach, where Epi-Curriculum outperforms the baseline on unseen and seen domains by 2.28 and 3.64 on En-De task, and 3.32 and 2.23 on En-Ro task.
The results also demonstrate that the episodic framework is more effective than the MAML-based framework. 


\section*{Limitations}
Despite the robustness and adaptability, our Epi-Curriculum also has one essential limitation.
The episodic framework has a high computational cost in both time complexity and space complexity.
\textbf{Time Complexity:}
If we assume O(1) is the time complexity for updating a model's parameters once in each iteration, such as the conventional AGG.
However, the episodic framework begins with O(N) for training N source domains once, followed by O(2) for applying episodic encoder training and episodic decoder training, and last comes with O(1) for the final update combined with episodic encoder loss and episodic decoder loss.
In the experiment of both En-De and En-Ro tasks, there are 5 domains in training and the training time of the episodic framework is eight times that of AGG.
In practice, AGG needs only 15 minutes to train an epoch on the En-De task but the episodic framework requires approximately 2 hours.
Apparently, the additional training time can not be ignored with an increasing number of source domains.
\textbf{Space Complexity:}
Let O(1) be the space complexity for storing the parameters of one model, such as one single AGG.
Whereas our episodic framework requires O(N) space to store N domain-specific models and O(1) space to store the AGG model.
The additional space requirement also can not be ignored with a larger translation model.

\section*{Ethics Statement}
Carefully reviewing the ACL Code of Ethics, we consider this work does not have ethical issues, including the use of data, and the potential applications of our work.

\bibliography{anthology,custom}
\bibliographystyle{acl_natbib}

\appendix
\section{Data Statistics and Preprocessing}
\label{appendix_data_sta}

\begin{table}[t]
\resizebox{\columnwidth}{!}{
\centering
\begin{tabular}{c c ccc}
    \hline
    & \multirow{2}{*}{\textbf{Domain}} & \textbf{Training} & \textbf{Fine-Tuning} & \textbf{Testing} \\
    &   & \textbf{(960k)} & \textbf{(10k)} & \textbf{(20k)} \\
    \hline
     \multirow{5}{*}{Unseen} & \textit{Covid-19} &  / & 338 & 714 \\
    & \textit{Bible} & / & 300 & 640 \\
    & \textit{Books} &  / & 303 & 723 \\
    & \textit{ECB} &   / & 298 & 704 \\
    & \textit{TED2013} &   / & 417 & 939 \\
    \hline
    \multirow{5}{*}{Seen} & \textit{EMEA} &  33,067 & 610 & 1,315 \\
    & \textit{Tanzil} &   43,779 & 476 & 1,033 \\
    & \textit{KDE4} &   75,610 & 813 & 1,794 \\
    & \textit{OpenSub} &   86,499 & 1,118 & 2,624 \\
    & \textit{JRC} &   29,071 & 312 & 675 \\
    \hline
\end{tabular}
}
\caption{The sampled amount of sentences used in Training, Fine-Tuning, and Testing for the En-De task.}
\label{en_de_data_split}
\end{table}

\begin{table}[t]
\resizebox{\columnwidth}{!}{
\centering
\begin{tabular}{c c ccc}
    \hline
    & \multirow{2}{*}{\textbf{Domain}} & \textbf{Training} & \textbf{Fine-Tuning} & \textbf{Testing} \\
    &   & \textbf{(960k)} & \textbf{(10k)} & \textbf{(20k)} \\
    \hline
     \multirow{5}{*}{Unseen} & \textit{KDE4} & / & 1,006 & 2,190 \\
    & \textit{Bible} &  / & 301 & 625 \\
    & \textit{QED} &   / & 566 & 1,160 \\
    & \textit{GlobalVoices} &   / & 417 & 790 \\
    \hline
    \multirow{5}{*}{Seen} & \textit{EMEA} &  43,779 & 509 & 1,085 \\
    & \textit{Tanzil} &   38,821 & 461 & 850 \\
    & \textit{TED2013} &   42,327 & 477 & 1,015 \\
    & \textit{OpenSub} &   88,257 & 1,022 & 2,013 \\
    & \textit{JRC} &   30,306 & 338 & 724 \\
    \hline
\end{tabular}
}
\caption{The sampled amount of sentences used in Training, Fine-Tuning, and Testing for the En-Ro task.}
\label{en_ro_data_split}
\end{table}

Table \ref{en_de_data_split} and Table \ref{en_ro_data_split} present data statistics for the English-German (En-De) and English-Romanian (En-Ro) tasks, respectively.
The number of tokens is fixed due to the variety of each domain's average sentence length, where the number of Fine-Tuning is small to simulate the low-resource scenario. 
Following \cite{sharaf-etal-2020-meta}, the number of tokens in each set is 960k tokens in the Training set, 10k tokens in the Fine-Tuning set, and 20k tokens in the Testing set.
All the corpora are processed by sentencepiece\footnote{\url{https://github.com/google/sentencepiece}} with the vocabulary size of 32,128. 
We filter the length of sentences and keep the sentences no longer than 175 and no shorter than 5. 
Because long sentences require larger computational space and short sentences are too easy to translate, such as ``Yes, I am'' and ``You are welcome''.
\section{Implementation Details}
\label{appendix_training_details}

\begin{figure}[t]
\centering
\includegraphics[width=1\columnwidth]{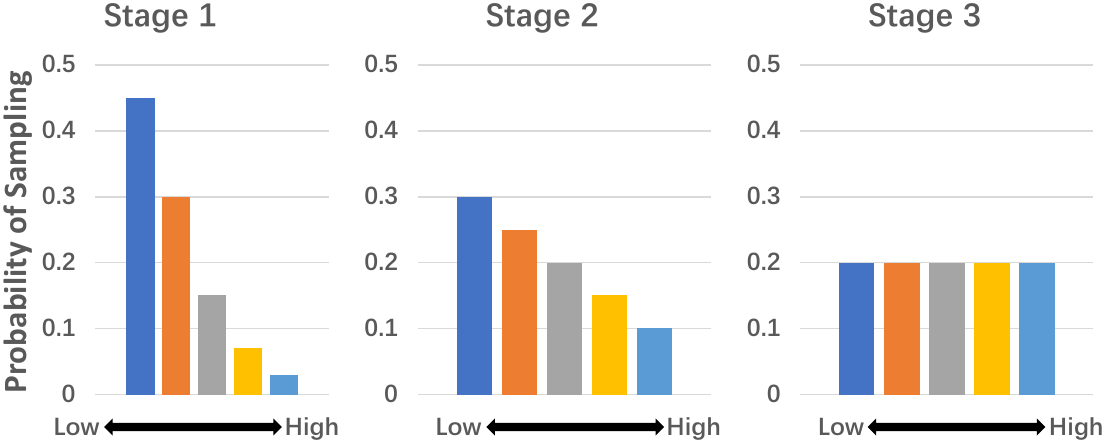} 
\caption{Probabilistic view of the \textbf{Default} training scheduler. The samples are arranged from low to high divergence scores on the x-axis. The y-axis is the shared probability of sampling the samples in each shard.}
\label{probability-sample}
\end{figure}

\begin{table*}[t]
\centering
\resizebox{2.05\columnwidth}{!}{
\begin{tabular}{c c cccc c ccccc}
\hline
&  & \multicolumn{4}{c}{\textbf{Unseen}} & &  \multicolumn{5}{c}{\textbf{Seen (Training Domains)}} \\
\cmidrule{3 - 6} \cmidrule{8 - 12}
& & \textit{KDE4} & \textit{Bible} & \textit{QED} & \textit{GlobalVoices} & &  \textit{EMEA} & \textit{Tanzil} & \textit{TED2013} & \textit{OpenSub} & \textit{JRC}\\
\hline
 & Vanilla & 24.99 & 7.61 & 22.87 & 24.08 &  & 31.90 & 5.57 & 25.75 & 16.10 & 36.09 \\
& AGG & 27.69 & \textbf{9.02} & 24.64 & 25.22 &  & 49.31 & 15.25 & 28.15 & 21.17 & 48.34  \\
Before & AGG-Curriculum & \textbf{28.29} & 8.95 & 24.68 & 25.08 &  & 50.07 & 17.54 & 27.88 & 20.35 & 47.42 \\
FT & Meta-MT & 27.82 & 8.59 & 24.70 & 25.16 &  & 48.69 & 14.95 & 28.48 & 20.63 & 47.76  \\
& Epi-NMT & 27.53 & 8.60 & \textbf{25.69} & \textbf{25.52} & & \textbf{53.38} & \textbf{20.75} & 29.79 & 21.71 & 49.57 \\
& Epi-Curriculum & 27.27 & 8.63 & 25.31 & 25.25 & & 53.24 & 18.72 & \textbf{30.16} & \textbf{21.79} & \textbf{49.97} \\
\hline

& Vanilla & 26.88 & 8.53 & 23.02 & 25.04 &  & 32.77 & 6.21 & 26.26 & 16.68 & 37.25 \\
& AGG & 29.26 & 10.08 & 24.92 & 26.23 &  & 49.86 & 15.49 & 29.15 & 21.41 & 48.86  \\
After & AGG-Curriculum & 30.27 & 11.02 & 25.02 & 27.19 & & 51.13 & 18.31 & 28.36 & 20.80 & 47.67 \\
FT& Meta-MT & 30.15 & 10.39 & 24.91 & 27.35 &  & 49.89 & 15.76 & 28.87 & 21.45 & 48.34 \\
& Epi-NMT & 31.22 & 10.82 & 25.76 & \textbf{28.27} & & \textbf{54.20} & \textbf{21.59} & 29.80 & 22.14 & 49.64 \\
& Epi-Curriculum & \textbf{31.45} & \textbf{11.12} & \textbf{25.94} & 28.25 &  & 53.86 & 19.42 & \textbf{30.21} & \textbf{22.52} & \textbf{49.91} \\
\hline

\multirow{6}{*}{$\triangle$FT} & Vanilla & 1.89 & 0.92 & 0.15 & 0.96 &  & 0.87 & 0.64 & 0.51 & 0.58 & \textbf{1.16} \\
& AGG & 1.57 & 1.06 & 0.28 & 1.01 & & 0.55 & 0.24 & 0.21 & 0.24 & 0.52 \\
& AGG-Curriculum & 1.98 & 2.07 & 0.34 & 2.11 & & 1.05 & 0.77 & \textbf{0.56} & 0.45 & 0.25 \\
& Meta-MT & 2.33 & 1.80 & 0.21 & 2.19 & & \textbf{1.20} & 0.81 & 0.39 & \textbf{0.82} & 0.57  \\
& Epi-NMT & 3.69 & 2.22 & 0.07 & 2.75 &  & 0.82 & \textbf{0.84} & 0.01 & 0.43 & 0.07 \\
& Epi-Curriculum & \textbf{4.18} & \textbf{2.49} & \textbf{0.63} & \textbf{3.00} &  & 0.62 & 0.70 & 0.05 & 0.73  & -0.06 \\
\hline
\end{tabular}
}
\caption{BLEU scores over Testing sets on En-Ro task. Before/After FT denotes the performance before/after individual fine-tuning. $\triangle$FT denotes the improvement of fine-tuning. The best results are highlighted in bold.}
\label{table_ro}
\end{table*}

\subsection{Hyperparameters}
\label{appendix_hyperparameters}
To follow the setting in T5, all experiments are trained using Adafactor \cite{shazeer2018adafactor} optimizer but with \textit{scale$\_$parameter = False} and \textit{relative$\_$step = False} to keep a consistent learning rate.
For the hyperparameters, we use $\alpha$ = 3e-5 and $\beta$ = 5e-5.
Our approach can be converged within 9 epochs and all the experiments are run on an NVIDIA RTX 3090 GPU with 24 GB of memory.

\subsection{Curriculum Training Scheduler}
\label{appendix_curriculum_training_scheduler}
We follow the general difficulty measure (domain divergence measure in our case) and training scheduler to conduct our curriculum learning.
Each training sample is first evaluated by Equation \ref{denoise_equation} and is filtered if the sample has negative $q_Z$ (mentioned in section \ref{sec_denoise}).
There are approximately 8\% of the total training samples being filtered for En-De and En-Ro tasks.
The rest samples are then scored by Equation \ref{divergence_score} and sorted in ascending order.
We evenly divide the sorted data into 5 shards, such that their average scores are from low to high.
To implement the training scheduler that begins with low-divergence samples, we start with higher probabilities for low-divergence samples to be sampled.
Gradually, in the later training stages, we increase the probabilities for high-divergence samples.
At the final stage, all samples have an equal probability of being sampled.
Figure \ref{probability-sample} illustrates this default probabilistic view, which indicates the probability of sampling in the three stages.


\section{Results and Discussion (En-Ro)}
\label{appendix_en_ro}

\begin{table}[t]
\centering
\resizebox{1\columnwidth}{!}{
\begin{tabular}{c | c c | c c }
    \hline
    & \multicolumn{2}{c}{\textbf{Before FT}} & \multicolumn{2}{c}{\textbf{After FT ($\triangle$FT)}}\\
    \hline
     & \textbf{Unseen} & \textbf{Seen} & \textbf{Unseen} & \textbf{Seen} \\
    \hline
    AGG & 1.76 & / & 1.76 (0.09) & / (/)\\
    AGG-Curriculum & 1.86 & 0.21 & 2.51 (0.54) & 0.30 (0.09)\\
    Meta-MT & 1.68 & -0.34 & 2.33 (0.50) & -0.09 (\textbf{0.25})\\
    Epi-NMT & \textbf{1.95} & \textbf{2.60} & 3.15 (0.70)  & \textbf{2.52} (-0.08)\\
    Epi-Curriculum & 1.73 & 2.33 & \textbf{3.32 (0.91)} & 2.23 (-0.10)\\
    \hline
\end{tabular}
}
\caption{The average improvement of Before FT, After FT, and $\triangle$FT over baselines (Vanilla for unseen and AGG for seen) on the En-Ro task.}
\label{en_ro_small}
\end{table}

The results of English-Romanian (En-Ro) are shown in Table \ref{table_ro} and Table \ref{en_ro_small}.
\textbf{Robustness:}
From the results Before FT in Table \ref{table_ro}, we can observe that:
(i) The meta-learning approach Meta-MT is worse than the AGG in 6 out of 9 domains. 
(ii) Epi-NMT outperforms AGG-Curriculum in 6 out of 9 domains.
(iii) Our episodic-based approaches (Epi-NMT and Epi-Curriculum) have the best performance in 7 out of 9 domains, especially the strength in seen domains.

\textbf{Adaptability:}
Based on the results of After FT and $\triangle$FT in Table \ref{table_ro}, we can observe that:
(i) Meta-MT also shows the same pattern that surpasses the AGG after fine-tuning in 6 domains. 
(ii) Our episodic-based approaches (Epi-NMT and Epi-Curriculum) monopolize all the scores after fine-tuning, where the Epi-Curriculum performs the best in 6 out of 9.
(iii) Epi-NMT and Epi-Curriculum also show their superiority in $\triangle$FT, where the Epi-Curriculum achieves the greatest improvement in all the unseen domains.

\textbf{Summary:}
From the results in Table \ref{en_ro_small}:
(i) Meta-MT still has strength in adaptability (0.50 and 1.15) and weakness in robustness (1.68 and -0.34).
(ii) Epi-NMT and Epi-Curriculum outperform the other approaches in most cases except the $\triangle$FT in seen domains (-0.08 and -0.1), where only AGG-Curriculum (0.09) and Meta-MT (0.25) have positive improvement compared to the baseline AGG. 
(iii) Epi-NMT outperforms Epi-Curriculum in 3 out of 4 cases, only worse than Epi-Curriculum in unseen domains After FT. 
But Epi-Curriculum performs better than Epi-NMT in more domains in Table \ref{table_ro}.
The main advantage of Epi-NMT originates from the strength in the Tanzil Domain (20.75 vs 18.72 and 21.59 vs 19.42).

\section{Robustness to Parameter Perturbation}
\label{appendix_perturbation}
\begin{figure*}[t]
\centering
\includegraphics[width=2.05\columnwidth]{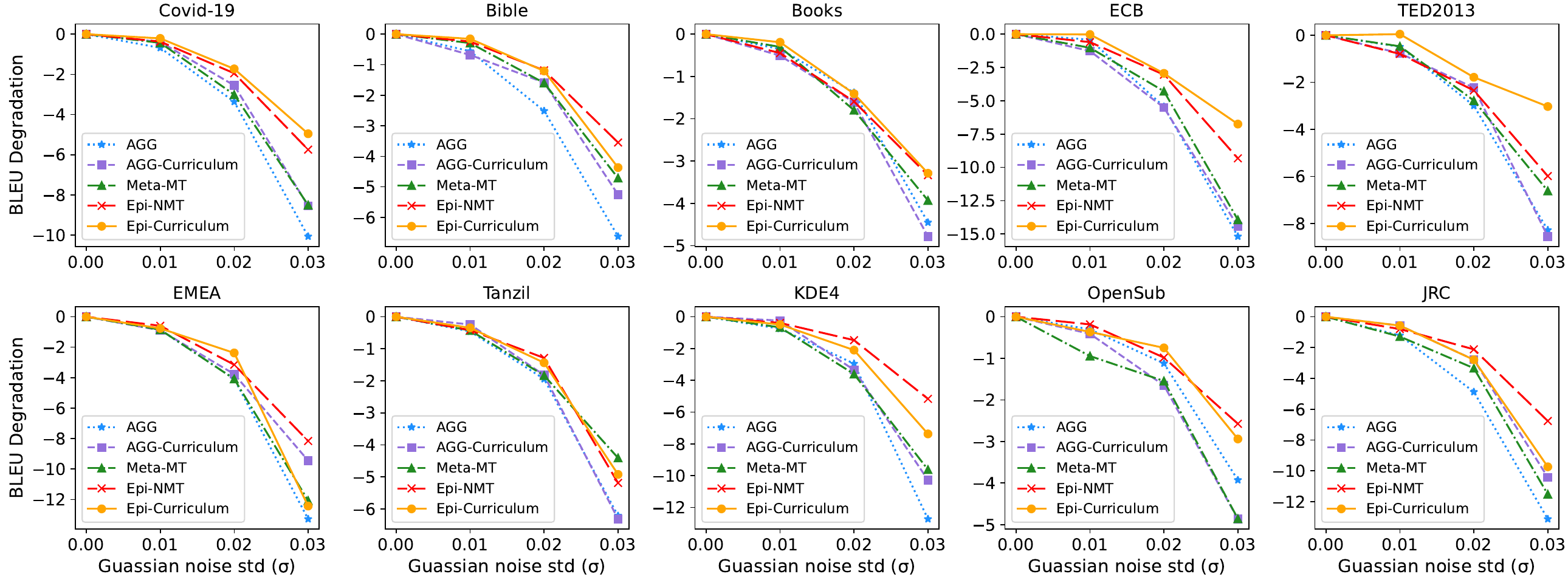} 
\caption{BLEU score degradation after adding Gaussian noise with different standard deviations ($\sigma$) to the models trained by AGG, AGG-Curriculum, Meta-MT, Epi-NMT, and Epi-Curriculum. Best viewed in color.}
\label{sharpness}
\end{figure*}

In terms of a model's robustness, recent studies have analyzed the quality of the minima that the model falls into \cite{keskar2017large, zhang2018deep, pmlr-v162-qu22a}.
Rather than having parameters that only themselves have low loss values, it suggests that a model with good generalization ability should seek parameters that lie in neighborhoods having uniformly low loss values.
In other words, a robust model will be obtained by converging to a flat minimum, instead of a sharp one.
Therefore, if a model's performance is not dependent on a precisely tuned solution, it would less likely to suffer from parameter perturbations.

To this end, it is natural to compare the robustness of our approach and others by simulating the parameter perturbations.
In detail, we observe the translation performance degradation by increasingly adding Gaussian noise to the model's parameters of the En-De task.
Results are shown in Figure \ref{sharpness}.
We can observe that:
(i) BLEU scores decrease as the parameters are perturbed harder.
(ii) Although the performance is similar when the perturbation is not much (0.01, 0.02), it is obvious that the models trained by our episodic framework (Epi-NMT and Epi-Curriculum) drop the slowest.
(iii) When we have the biggest perturbation ($\sigma$ = 0.03), Epi-NMT and Epi-Curriculum have the best performance in 9 out of 10 domains, only slightly worse than Meta-MT in Tanzil.

This experiment suggests that the minima found by our episodic framework have higher quality.
The minima drop into a wider area that neighbor points are also with low loss values, and further demonstrate the robustness of the episodic framework.

\section{Curriculum Validity (En-Ro)}
\label{appendix_curr_divergence}
\begin{figure}[t]
\centering
\includegraphics[width=0.75\columnwidth]{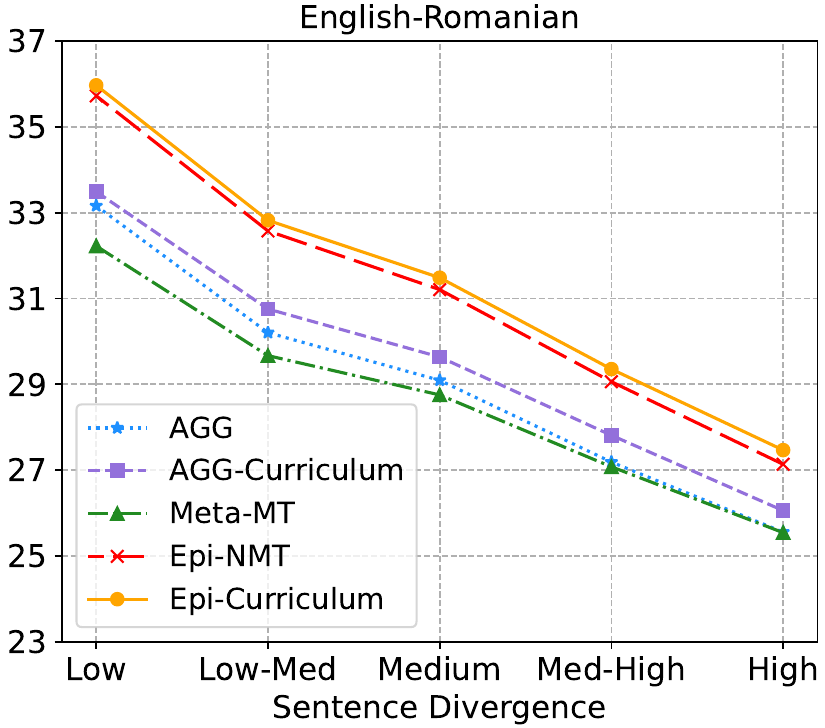} 
\caption{Average performance over Testing sets across five levels of divergence on the seen domains (En-Ro).}
\label{curr_ro}
\end{figure}

The results of curriculum validity on the En-Ro task are shown in Figure \ref{curr_ro}.
Epi-NMT and Epi-Curriculum also perform the best at all levels, and Epi-Curriculum has very slight but consistent strength over Epi-NMT with increasing divergence.

\section{Impact of Training Scheduler}
\label{appendix_impact_scheduler}

\begin{table*}[t]
\centering
\resizebox{2.05\columnwidth}{!}{
\begin{tabular}{c c ccccc c ccccc}
\hline
&  & \multicolumn{5}{c}{\textbf{Unseen}} & &  \multicolumn{5}{c}{\textbf{Seen (Training Domains)}} \\
\cmidrule{3 - 7} \cmidrule{9 - 13}
& &  \textit{Covid-19} & \textit{Bible} & \textit{Books} & \textit{ECB} & \textit{TED2013} & &  \textit{EMEA} & \textit{Tanzil} & \textit{KDE4} & \textit{OpenSub} & \textit{JRC} \\
\hline
& & \multicolumn{10}{c}{\textbf{AGG-Curriculum}}\\
\hline
& Default & 26.05 & 12.06 & \textbf{11.36} & 32.66 & \textbf{26.91} & &  38.46 & 15.89 & 30.94 & 18.73 & \textbf{40.61} \\
Before FT& Advanced & \textbf{26.31} & \textbf{12.55} & 11.24 & \textbf{32.81} & 26.78 & & 39.11 & \textbf{17.59} & \textbf{31.50} & \textbf{19.04} & 40.19 \\
& Reversed & 25.64 & 12.38 & 10.83 & 32.27 & 26.66 & & \textbf{39.26} & 15.27 & 31.46 & 18.56 & 39.55  \\
\hline
& Default & \textbf{26.71} & 13.48 & 11.77 & \textbf{33.76} & \textbf{28.36} & & 39.40 & 16.78 & 31.24 & 19.43 & \textbf{41.10} \\
After FT& Advanced & 26.57 & \textbf{13.56} & \textbf{11.97} & 33.53 & 28.20 & & \textbf{40.37} & \textbf{18.24} & 31.94 & \textbf{19.54} & 40.76  \\
& Reversed & 26.18 & 13.14 & 11.39 & 33.46 & 27.51 & & 40.16 & 16.05 & \textbf{32.08} & 18.83 & 40.45  \\
\hline
& & \multicolumn{10}{c}{\textbf{Epi-Curriculum}}\\
\hline

& Default & \textbf{27.12} & \textbf{12.70} & \textbf{11.90} & 34.12 & \textbf{26.63} & &  43.51 & 18.72 & 32.87 & \textbf{20.44} & 42.30  \\
Before FT& Advanced & 26.43 & 12.31 & 11.87 & 34.01 & 26.42 & & 44.56 & \textbf{20.28} & 33.25 & 20.22 & \textbf{43.14}  \\
& Reversed & 26.44 & 12.19 & 11.65 & \textbf{34.28} & 26.17 & & \textbf{44.71} & 17.78 & \textbf{33.27} & 20.17 & 43.06  \\
\hline
& Default & \textbf{27.73} & \textbf{15.11} & \textbf{12.53} & 34.89 & \textbf{29.11} & & 44.18 & 20.62 & 33.39 & \textbf{20.89} & 43.24 \\
After FT& Advanced & 27.07 & 14.51 & 12.48 & \textbf{34.93} & 28.77 & & 44.74 & \textbf{21.30} & \textbf{33.53} & 20.65 & \textbf{43.46}  \\
& Reversed & 27.13 & 14.40 & 12.34 & 34.73 & 28.82 & & \textbf{44.79} & 19.66 & 33.18 & 20.38 & 43.23  \\
\hline
\end{tabular}
}
\caption{BLEU scores over Testing sets on En-De task with different training schedulers. Before/After FT denotes the performance before/after individual fine-tuning. The best results are highlighted in bold.}
\label{training_scheduler_de}
\end{table*}

\begin{table*}[t]
\centering
\resizebox{2.05\columnwidth}{!}{
\begin{tabular}{c c cccc c ccccc}
\hline
&  & \multicolumn{4}{c}{\textbf{Unseen}} & &  \multicolumn{5}{c}{\textbf{Seen (Training Domains)}} \\
\cmidrule{3 - 6} \cmidrule{8 - 12}
& & \textit{KDE4} & \textit{Bible} & \textit{QED} & \textit{GlobalVoices} & &  \textit{EMEA} & \textit{Tanzil} & \textit{TED2013} & \textit{OpenSub} & \textit{JRC}\\
\hline
& & \multicolumn{9}{c}{\textbf{AGG-Curriculum}}\\
\hline
& Default & 28.29 & \textbf{8.95} & 24.68 & 25.08 &  & 50.07 & 17.54 & 27.88 & 20.35 & 47.42 \\
Before FT& Advanced & \textbf{28.42} & 8.71 & 24.63 & 25.29 &  & \textbf{51.47} & \textbf{18.11} & 28.16 & 20.30 & 47.76   \\
& Reversed & 28.14 & 8.61 & \textbf{24.80} & \textbf{25.78} &  & 49.37 & 14.81 & \textbf{28.42} & \textbf{20.58} & \textbf{48.18}  \\
\hline
& Default & 30.27 & 10.72 & 25.02 & \textbf{27.19} & & 51.13 & 18.31 & 28.36 & 20.80 & 47.67 \\
After FT& Advanced & \textbf{30.38} & \textbf{10.75} & 24.86 & 26.47 &  & \textbf{51.91} & \textbf{19.16} & \textbf{28.80} & 21.12 & 48.10  \\
& Reversed & 30.32 & 10.63 & \textbf{25.31} & 27.18 &  & 50.40 & 15.64 & 28.67 & \textbf{21.41} & \textbf{48.33}  \\
\hline
& & \multicolumn{9}{c}{\textbf{Epi-Curriculum}}\\
\hline

& Default & 27.27 & 8.63 & 25.31 & 25.25 & & 53.24 & 18.72 & 30.16 & \textbf{21.79} & 49.97 \\
Before FT& Advanced & 27.23 & \textbf{8.79} & 25.42 & 25.06 &  & \textbf{54.27} & \textbf{22.21} & 29.76 & 21.73 & 49.71  \\
& Reversed & \textbf{27.41} & 8.77 & \textbf{26.06} & \textbf{26.06} &  & 52.43 & 16.76 & \textbf{30.39} & 21.57 & \textbf{50.00}  \\
\hline
& Default & \textbf{31.45} & \textbf{11.12} & 25.94 & 28.25 &  & 53.86 & 19.42 & 30.21 & \textbf{22.52} & 49.91 \\
After FT & Advanced & 30.70 & 10.77 & 25.48 & 28.14 &  & \textbf{55.18} & \textbf{22.77} & 29.93 & 21.97 & 49.88  \\
& Reversed & 31.03 & 10.81 & \textbf{26.19} & \textbf{28.43} &  & 53.47 & 18.03 & \textbf{30.29} & 22.23 & \textbf{49.97}  \\
\hline
\end{tabular}
}
\caption{BLEU scores over Testing sets on En-Ro task with different training schedulers. Before/After FT denotes the performance before/after individual fine-tuning. The best results are highlighted in bold.}
\label{training_scheduler_ro}
\end{table*}

\begin{figure}[t]
\centering
\includegraphics[width=1\columnwidth]{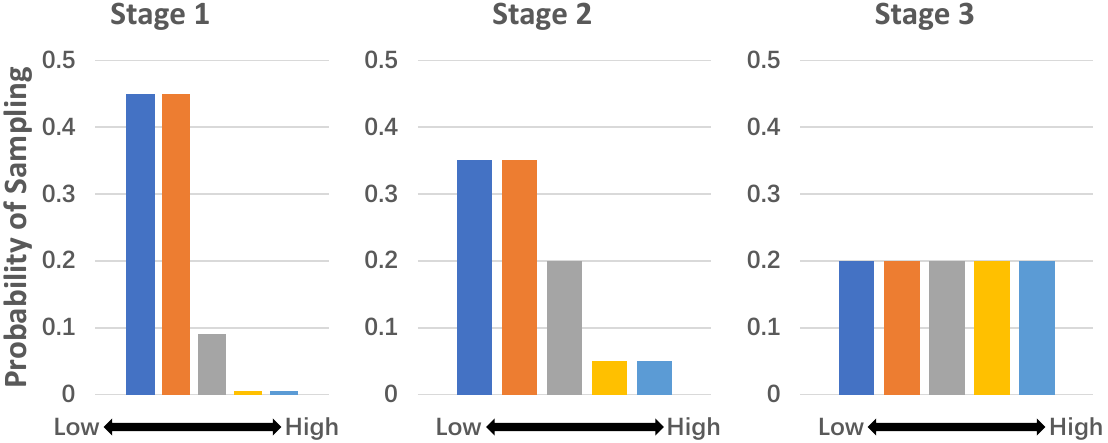} 
\caption{Probabilistic view of the \textbf{Advanced} training scheduler. The samples are arranged from low to high divergence scores on the x-axis. The y-axis is the shared probability of sampling the samples in each shard.}
\label{probability-sample-advanced}
\end{figure}

\begin{figure}[t]
\centering
\includegraphics[width=1\columnwidth]{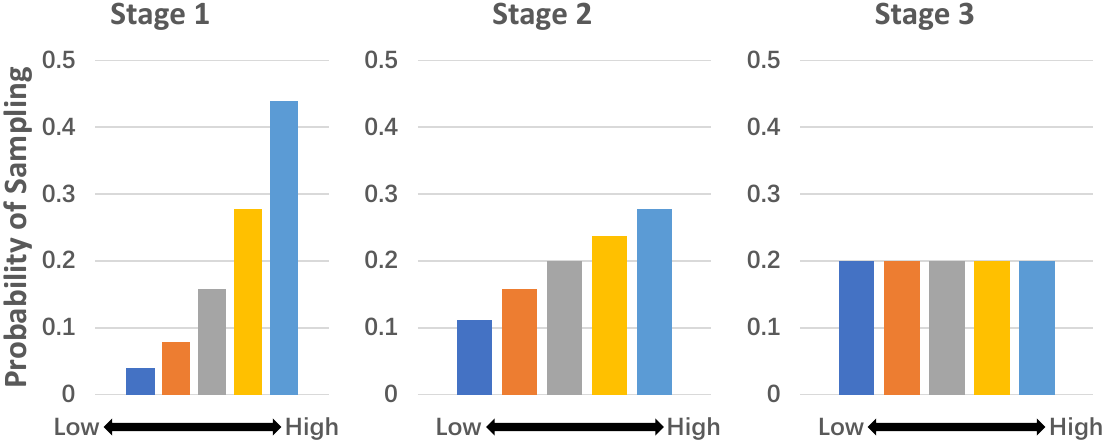} 
\caption{Probabilistic view of the \textbf{Reversed} training scheduler. The samples are arranged from low to high divergence scores on the x-axis. The y-axis is the shared probability of sampling the samples in each shard.}
\label{probability-sample-reversed}
\end{figure}

We introduce another two training schedulers to compare with our default scheduler (describe in Appendix \ref{appendix_curriculum_training_scheduler}):
\begin{itemize}
    \item \textbf{Advanced:} The shards with low divergence scores have more probabilities to be sampled in the first two stages, and equal probability in the last stage. Figure \ref{probability-sample-advanced} illustrates this probability view.
    \item \textbf{Reversed:}: The shards are sorted in descending order of divergence. Figure \ref{probability-sample-reversed} shows this probability view.
\end{itemize}

Table \ref{training_scheduler_de} and Table \ref{training_scheduler_ro} show the results with different training schedulers on En-De and En-Ro tasks, respectively.
We can see that different training schedulers do not lead to significant performance changes except Tanzil.
Tanzil is sensitive to different training schedulers, where its performance can be summarized as Reversed < Default < Advanced.
Since this order is also related to the portion that low divergence samples are sampled during the training process.
Thus the performance of Tanzil may depend on the performance of the low-divergence corpus.


\section{Impact of Denoising}
\label{appendix_denoise_impact}

\begin{table*}[t]
\centering
\resizebox{2.05\columnwidth}{!}{
\begin{tabular}{c c ccccc c ccccc}
\hline
&  & \multicolumn{5}{c}{\textbf{Unseen}} & &  \multicolumn{5}{c}{\textbf{Seen (Training Domains)}} \\
\cmidrule{3 - 7} \cmidrule{9 - 13}
& &  \textit{Covid-19} & \textit{Bible} & \textit{Books} & \textit{ECB} & \textit{TED2013} & &  \textit{EMEA} & \textit{Tanzil} & \textit{KDE4} & \textit{OpenSub} & \textit{JRC} \\
\hline
& & \multicolumn{10}{c}{\textbf{AGG-Curriculum}}\\
\hline
Before & Denoised & 26.05 & 12.06 & \textbf{11.36} & \textbf{32.66} & 26.91 & &  38.46 & \textbf{15.89} & 30.94 & 18.73 & \textbf{40.61} \\
FT& Noised & \textbf{26.27} & \textbf{12.61} & 10.89 & 32.30 & \textbf{27.73} & & \textbf{38.65} & 15.18 & \textbf{31.27} & \textbf{18.77} & 40.42 \\
\hline
Before & Denoised & 26.71 & 13.48 & \textbf{11.77} & \textbf{33.76} & 28.36 & & 39.40 & 16.78 & 31.24 & \textbf{19.43} & \textbf{41.10} \\
FT& Noised & \textbf{26.74} & \textbf{13.65} & 11.48 & 33.56 & \textbf{28.85} & & \textbf{40.05} & \textbf{17.10} & \textbf{31.74} & 19.17 & 40.67  \\
\hline
& & \multicolumn{10}{c}{\textbf{Epi-Curriculum}}\\
\hline

Before & Denoised & \textbf{27.12} & \textbf{12.70} & 11.90 & \textbf{34.12} & 26.63 & &  \textbf{43.51} & 18.72 & 32.87 & \textbf{20.44} & \textbf{42.30}  \\
FT& Noised & 27.03 & 12.38 & \textbf{12.16} & 34.07 & \textbf{27.18} & & 42.87 & \textbf{19.26} & \textbf{32.96} & 19.87 & 42.24  \\
\hline
After & Denoised & \textbf{27.73} & \textbf{15.11} & \textbf{12.53} & \textbf{34.89} & \textbf{29.11} & & \textbf{44.18} & \textbf{20.62} & 33.39 & \textbf{20.89} & 43.24 \\
FT& Noised & 27.69 & 14.42 & 12.51 & 34.26 & 28.91 & & 43.92 & 20.48 & \textbf{33.59} & 20.00 & \textbf{43.39}  \\
\hline
\end{tabular}
}
\caption{BLEU scores over Testing sets on En-De task with and without noise data. Before/After FT denotes the performance before/after individual fine-tuning. The best results are highlighted in bold.}
\label{curriculum_denoise_de}
\end{table*}

\begin{table*}[t]
\centering
\resizebox{2.05\columnwidth}{!}{
\begin{tabular}{c c cccc c ccccc}
\hline
&  & \multicolumn{4}{c}{\textbf{Unseen}} & &  \multicolumn{5}{c}{\textbf{Seen (Training Domains)}} \\
\cmidrule{3 - 6} \cmidrule{8 - 12}
& & \textit{KDE4} & \textit{Bible} & \textit{QED} & \textit{GlobalVoices} & &  \textit{EMEA} & \textit{Tanzil} & \textit{TED2013} & \textit{OpenSub} & \textit{JRC}\\
\hline
& & \multicolumn{9}{c}{\textbf{AGG-Curriculum}}\\
\hline
Before & Denoised & \textbf{28.29} & 8.95 & \textbf{24.68} & 25.08 &  & 50.07 & 17.54 & \textbf{27.88} & 20.35 & 47.42 \\
FT & Noised & 28.25 & \textbf{9.05} & 24.37 & \textbf{25.18} &  & \textbf{50.43} & \textbf{18.28} & 27.35 & \textbf{20.56} & \textbf{47.78}  \\
\hline
After& Denoised & \textbf{30.27} & 10.72 & 25.02 & \textbf{27.19} & & 51.13 & 18.31 & 28.36 & \textbf{20.80} & 47.67 \\
FT & Noised & 30.21 & \textbf{11.03} & \textbf{25.07} & 26.08 &  & \textbf{51.52} & \textbf{19.26} & \textbf{28.77} & 20.62 & \textbf{47.91}  \\
\hline
& & \multicolumn{9}{c}{\textbf{Epi-Curriculum}}\\
\hline

Before & Denoised & \textbf{27.27} & 8.63 & 25.31 & \textbf{25.25} & & \textbf{53.24} & 18.72 & \textbf{30.16} & 21.79 & \textbf{49.97} \\
FT& Noised & 27.06 & \textbf{8.65} & \textbf{25.56} & 24.99 &  & 51.87 & \textbf{18.95} & 29.69 & \textbf{22.15} & 48.89  \\
\hline
After & Denoised & 31.45 & \textbf{11.12} & \textbf{25.94} & \textbf{28.25} &  & \textbf{53.86} & 19.42 & \textbf{30.21} & 22.52 & \textbf{49.91} \\
FT & Noised & \textbf{31.87} & 10.78 & 25.72 & 27.76 &  & 52.86 & \textbf{19.66} & 29.81 & \textbf{22.85} & 49.07  \\
\hline
\end{tabular}
}
\caption{BLEU scores over Testing sets on En-Ro task with and without noise data. Before/After FT denotes the performance before/after individual fine-tuning. The best results are highlighted in bold.}
\label{curriculum_denoise_ro}
\end{table*}

Table \ref{curriculum_denoise_de} and Table \ref{curriculum_denoise_ro} show the results with and without noised data.
The results do not show a significant difference between the performance with noise and without noise on both En-De and En-Ro tasks.

\end{document}